\documentclass[sigconf]{acmart}

\copyrightyear{2021}
\acmYear{2021}
\setcopyright{acmcopyright}\acmConference[MM '21]{Proceedings of the 29th ACM International Conference on Multimedia}{October 20--24, 2021}{Virtual Event, China}
\acmBooktitle{Proceedings of the 29th ACM International Conference on Multimedia (MM '21), October 20--24, 2021, Virtual Event, China}
\acmPrice{15.00}
\acmDOI{10.1145/3474085.3475226}
\acmISBN{978-1-4503-8651-7/21/10}

\newcommand{\E}{\mathbb{E}}                   
\newcommand{\z}{{\rm\bf z}}                   
\newcommand{\w}{{\rm\bf w}}                   
\newcommand{\txt}{{\rm\bf t}}                   
\newcommand{\Z}{\mathcal{Z}}                  
\newcommand{\W}{\mathcal{W}}                  
\newcommand{\x}{{\rm\bf x}}                   
\newcommand{\X}{\mathcal{X}}                  
\newcommand{\Loss}{\mathcal{L}}               

\AtBeginDocument{%
  \providecommand\BibTeX{{%
    \normalfont B\kern-0.5em{\scshape i\kern-0.25em b}\kern-0.8em\TeX}}}

\begin{document}
\fancyhead{}

\title{Cycle-Consistent Inverse GAN for Text-to-Image Synthesis}

\author{Hao Wang$^{1,2}$, Guosheng Lin$^{1}$, Steven C. H. Hoi$^{3}$, Chunyan Miao$^{1,2}$}
\authornote{Corresponding author}
\affiliation{%
  \institution{$^{1}$School of Computer Science and Engineering, Nanyang Technological University (NTU), Singapore \\
  $^{2}$Joint NTU-UBC Research Centre of Excellence in Active Living for the Elderly, NTU, Singapore \\
  $^{3}$Singapore Management University, Singapore \\
  }
}
\email{{hao005,gslin,ascymiao}@ntu.edu.sg, chhoi@smu.edu.sg}

\renewcommand{\shortauthors}{Wang et al.}

\begin{abstract}
  This paper investigates an open research task of text-to-image synthesis for automatically generating or manipulating images from text descriptions. Prevailing methods mainly use the text as conditions for GAN generation, and train different models for the text-guided image generation and manipulation tasks. In this paper, we propose a novel unified framework of Cycle-consistent Inverse GAN (CI-GAN) for both text-to-image generation and text-guided image manipulation tasks. Specifically, we first train a GAN model without text input, aiming to generate images with high diversity and quality. Then we learn a GAN inversion model to convert the images back to the GAN latent space and obtain the inverted latent codes for each image, where we introduce the cycle-consistency training to learn more robust and consistent inverted latent codes. We further uncover the latent space semantics of the trained GAN model, by learning a similarity model between text representations and the latent codes. In the text-guided optimization module, we generate images with the desired semantic attributes by optimizing the inverted latent codes. Extensive experiments on the Recipe1M and CUB datasets validate the efficacy of our proposed framework.
\end{abstract}

\ccsdesc[500]{Computing methodologies~Computer vision}

\keywords{GAN, Text-to-image synthesis, Cycle consistency}

\maketitle

\begin{figure}
\begin{center}
\includegraphics[width=0.35\textwidth]{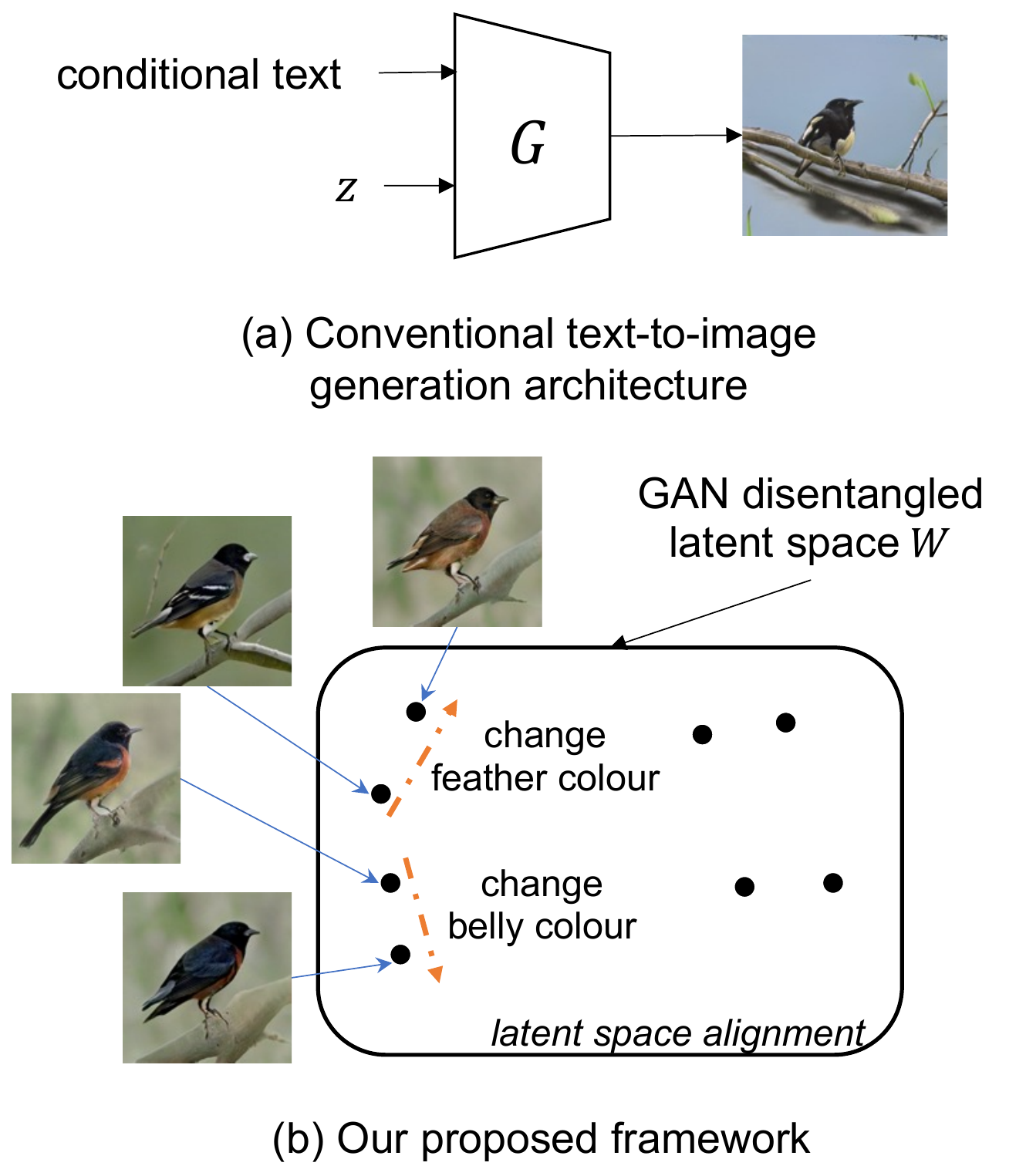}
\end{center}
\vspace{-5pt}
   \caption{The comparison of conventional architecture and our proposed framework for text-to-image generation. Existing works mainly take the text feature-conditioned GAN structure, where the limited combinations of text and images will affect the diversity of generation. While we adopt a decoupled learning scheme, we first train a GAN model without text, then we discover the semantics of the latent space $\W$ of the trained GAN. We allow the text representations to be matched with the latent codes, such that we can control the semantic attributes of the synthesised images by changing the latent codes.}
\vspace{-10pt}
\label{fig:demo}
\end{figure}

\section{Introduction}
Text-to-image synthesis \cite{zhang2018stackgan++,xu2018attngan,qiao2019mirrorgan,yin2019semantics,zhu2019dm,tao2020df,cheng2020rifegan,ramesh2021zero} aims to generate images that have semantic contents corresponding to the input text descriptions, typically based on the Generative Adversarial Networks (GANs) approaches. It has various potential applications, such as visual content design and art generation. However, text-to-image synthesis is a challenging cross-modal task, as we need to interpret the semantic attributes hidden in the text and produce images with high diversity and good quality. 

Prevailing works \cite{zhu2020cookgan,cheng2020rifegan,zhu2019dm,li2019controllable} on text-to-image generation mainly build their frameworks based on StackGAN \cite{zhang2018stackgan++}, which can generate high-resolution images progressively. Specifically, the StackGAN model stacks multiple generators and discriminators, which can generate initial low-resolution images with rough shapes and colour attributes first and then refine the initial images to the high-resolution ones. To improve the semantic correspondence between the textual descriptions and the generated images, Xu et al. propose AttnGAN \cite{xu2018attngan} to discover the attribute alignment between image and text by pretraining an attentional similarity model. However, the paired text-image training for GAN model limits the diversity of the model representation, since we only have limited combinations of text and images and the generated images are regularized by the corresponding real images and the text pairs. 
Moreover, it is hard to use the aforementioned framework to only change one attribute while preserve other text-irrelevant attributes in the generated images, hence we need to train an extra module to do the text-based image manipulation task \cite{li2020manigan}. 

The advent of style-based generator architecture, such as StyleGAN \cite{karras2019style,karras2020analyzing}, has greatly improved the realism, quality and diversity of the generated images. Specifically, the StyleGAN proposed to map the input noise to another latent space $\W$, which has been validated to yield more disentangled semantic representations. To uncover the relationships between the latent codes in the space $\W$ and the synthesised images, we need to be aware of the distributions of the space $\W$ and find the corresponding latent codes of the images. To this end, many research works adopt the GAN inversion technique \cite{zhu2020indomain,abdal2019image2stylegan,tov2021designing} to invert the images back to the space $\W$ and obtain the inverted latent codes. 

In this paper, we propose a novel framework of Cycle-consistent Inverse GAN (CI-GAN), where we incorporate the GAN inversion methodology to the text-to-image synthesis task. Technically, we first train a GAN inversion encoder to map the images to the latent space $\W$ of a trained StyleGAN, such that we can get the inverted latent codes for the real images of the given datasets. To make the original and inverted latent codes to be identical and follow the same distribution, we introduce to apply the cycle consistency loss on the GAN inversion training process, as obtaining similar inverted latent codes to the original ones is critical for our subsequent generation procedure.

We assume the StyleGAN learned space $\W$ is disentangled regarding the semantic attributes of the target image dataset. For example, in the bottom row of Figure \ref{fig:demo}, when we want to change the \emph{belly colour} of the bird image, the rest semantic attributes, such as the bird shape, pose and feather colour, will remain the same, only the bird \emph{belly colour} changes to the black colour from the orange colour. The disentanglement of the space $\W$ allows us to generate images with various attributes based on the optimization on the latent codes. To generate images from the textual descriptions, we learn a similarity model between text representations and the inverted latent codes, such that the latent codes can be optimized to have the desired semantic attributes. We feed the optimized latent codes into the trained StyleGAN generator and realize the text-to-image generation task. Apart from the text-to-image generation task, our proposed CI-GAN can also be used on the text-based image manipulation task by applying an extra perceptual loss between the original images and the images reconstructed from the optimized latent codes. 

Our contributions can be summarized as:
\begin{itemize}
   \item We propose a novel GAN approach combining GAN inversion and cycle consistency training for the text-to-image synthesis. The unified framework can be used for the text-to-image generation and text-based image manipulation tasks.
   \item We use the improved GAN inversion methods with cycle consistency training to invert real images to the GAN latent space and obtain the latent codes of images.
   \item We uncover the semantics of the latent codes, based on which we can generate high-quality images corresponding to the textual descriptions.
\end{itemize}
We evaluate our proposed framework CI-GAN on two public datasets in the wild, i.e. Recipe1M and CUB datasets. We conduct extensive experiments to analyse the efficacy of the CI-GAN. Finally, we present quantitative and qualitative results of our proposed methods and visualizations of the generated images.

\section{Related Work}

\subsection{Text-Based Image Generation}
In this section, we review two categories of text-based image generation, i.e. text-to-image generation and text-based image manipulation. Generating images from text is a challenging task, as we need to correlate the cross-modal information \cite{chen2020graph,wang2021cross,wang2020structure}. To control the correspondence between the text and the generated images, some prevailing text-to-image generation works \cite{cheng2020rifegan,zhu2019dm,li2019controllable} pretrain a Deep Attentional Multimodal Similarity Model (DAMSM) \cite{xu2018attngan}, which is used as a supervision to regularize the semantics of the generated images. Specifically, Cheng et al. \cite{cheng2020rifegan} propose to use a refinement module to return more complete caption sets, which can provide more semantic information for the image generation. Zhu et al. \cite{zhu2019dm} use a memory writing gate to refine the initial image and generate a high-quality one. Wang et al. \cite{wang2019learning} and Zhu et al. \cite{zhu2020cookgan} aim to generate food images from the cooking recipes.

For the text-based image manipulation task, it requires the model to only change certain parts or attributes and preserve other text-irrelevant attributes on the input images. Li et al. \cite{li2020manigan} propose a module to combine the text and generated images to jointly correlate the details, such that the mismatched attributes can be rectified. Dong et al. \cite{dong2017semantic} use an encoder-decoder architecture to take the original images as well as the textual descriptions as the input, and output the manipulated images, which is supervised by a discriminator.  

However, the existing text-to-image generation works suffer from the limited diversity of the generated images, since they use the paired text and images for the GAN training. Moreover, the aforementioned architectures adopt the multi-stage refinement \cite{zhang2018stackgan++,xu2018attngan} to improve the resolution of the generated images, therefore it is cumbersome to generate images with higher resolution. In contrast, our proposed method use the StyleGAN2 \cite{karras2020analyzing} model as the generator backbone and we do not use the paired text input when training GAN, which guarantees the quality and diversity of the generated images. 

\begin{figure*}
\begin{center}
\includegraphics[width=0.9\textwidth]{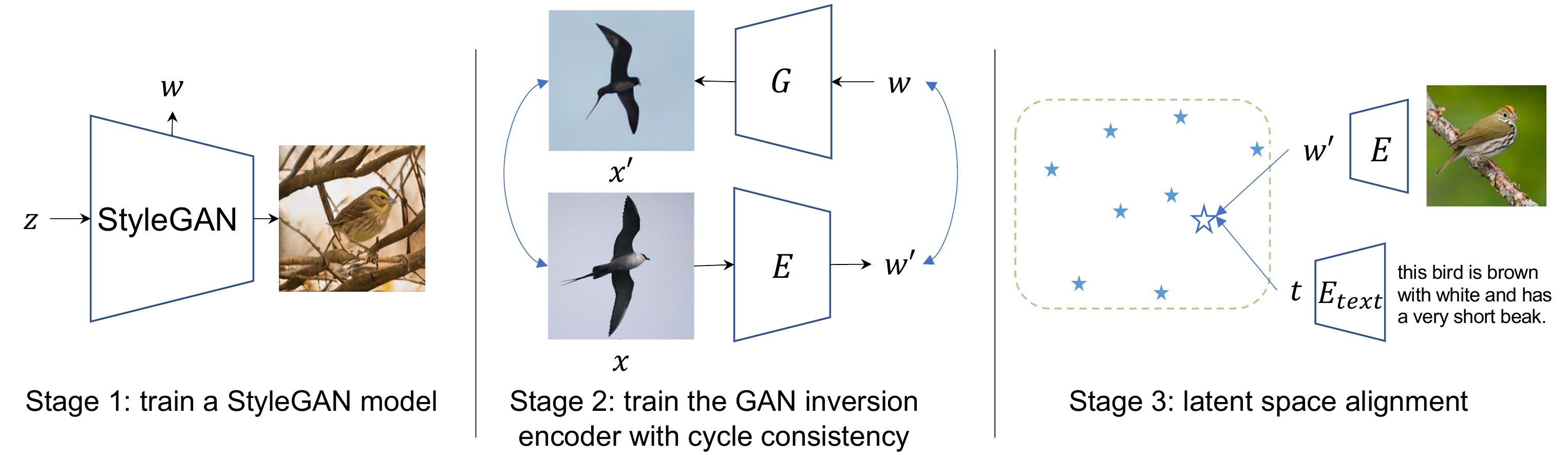}
\end{center}
\vspace{-10pt}
   \caption{The decoupled training flow of our proposed unified framework of Cycle-consistent Inverse GAN (CI-GAN). In Stage 1, we train a StyleGAN model without text input. The StyleGAN can map the input random noise $\z$ to the style latent codes $\w$, based on which the generator can produce images with high quality and diversity. In Stage 2, we train the GAN inversion encoder $E(\cdot)$ with the our proposed cycle-consistent learning, aiming to invert images $\x$ to the StyleGAN latent space $\W$ and get the inverted latent codes $\w'$. In Stage 3, we learn a latent space alignment model to train the text encoder $E_{text}(\cdot)$, such that the text extracted features $\txt$ can be mapped to the latent space $\W$ and aligned with the corresponding $\w'$.}
\vspace{-10pt}
\label{fig:framework}
\end{figure*}

\subsection{GAN Inversion}
Due to the lack of inference capabilities in GAN, the manipulation in the latent space can only be applied on the generated images, rather than any given real images. GAN inversion is popular way to manipulate the real images \cite{zhu2016generative,bau2020semantic,perarnau2016invertible}. The purpose of GAN inversion is to invert the given image to the latent space of a pretrained GAN model and obtain the inverted latent code, such that the image can be faithfully reconstructed by the generator from the inverted latent code. As a new technique to connect the images and the GAN latent space, GAN inversion enables the pretrained GAN model to be used in various image generation applications, such as image editing \cite{xia2021gan}.   

There are two main types of GAN inversion methods. The first one is learning-based GAN inversion method \cite{zhu2016generative,perarnau2016invertible}, which typically trains an encoder to generate the latent codes. The GAN generator will be fixed during training the encoder. The second method is optimization-based. This method typically reconstruct the original images by optimizing the latent codes, which is based on either gradient descent \cite{zhu2020indomain} or other iterative algorithms \cite{Karras2020ada}. Some works \cite{bau2019seeing,zhu2020indomain,xia2020tedigan} attempt to combine these two ideas, where the produced latent codes by the GAN inversion encoder are used as the initialization for the optimization step, as it is hard to get the perfect reconstructed images based on the inverted latent codes from a single GAN inversion encoder. Zhu et al. \cite{zhu2020indomain} propose to train the encoder based on the real images instead of the fake images, as a result, the trained encoder can be better adapted to the real scenarios. They also adopt the discriminator to compete with the GAN inversion encoder, in this way they can also involve semantic knowledge from the model. Abdal et al. \cite{abdal2020image2stylegan++} incorporate the spatial masks to learn both the input random noise and the latent codes in the StyleGAN latent space $\W$. Gu et al. \cite{gu2020image} employ multiple latent codes to generate feature maps at some intermediate layer of the generator, then compose them with adaptive channel importance to recover the input image. 

In our work, we combine the learning-based and optimization-based approaches to find the latent codes with the desired attributes. Existing works of GAN inversion mainly focus on the face image datasets \cite{zhu2020indomain, xia2020tedigan}, while we target two more challenging datasets in the wild, where the Recipe1M dataset contains cooked food images with multiple mixed ingredients and the CUB dataset has birds with various poses, shapes, colours and backgrounds. The difference between our proposed method and the existing works is that we use the cycle consistency training method, applying regularization on not only the image domain, but also the latent space domain. As a result, our framework can also get satisfactory inverted results in more complex scenarios compared to the face datasets, such as food and bird images in the wild.

\section{Method}
Our proposed Cycle-consistent Inverse GAN (CI-GAN) can be applied on both text-based image generation and manipulation tasks. The training pipeline is presented in Figure \ref{fig:framework}, which can be summarized in a three-stage training procedure:
\begin{itemize}
   \item \textbf{Stage 1.} We train a StyleGAN model without text input. The StyleGAN modal can map the random noise space $\Z$ to the style latent space $\W$, which has been proven to be more disentangled with various image properties.
   \item \textbf{Stage 2.} We propose to use the cycle-consistency training to learn the GAN inversion encoder, such that we can invert the real images and obtain corresponding latent codes $\w'$.
   \item \textbf{Stage 3.} We learn a latent space alignment model to align the text features $\txt$ with the corresponding inverted latent codes $\w'$, where we train the text encoder $E_{text}(\cdot)$.
\end{itemize}

\begin{figure}
\begin{center}
\includegraphics[width=0.45\textwidth]{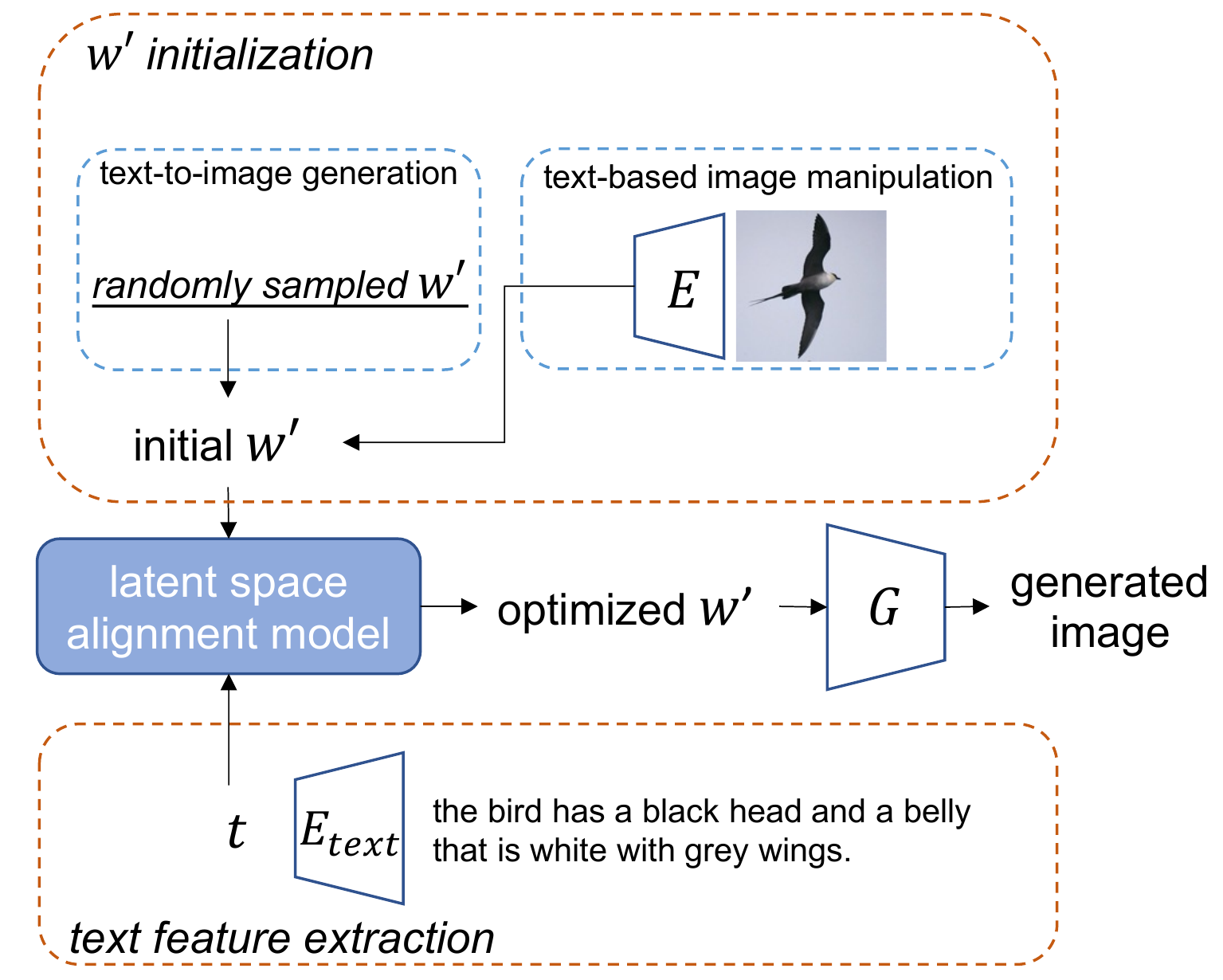}
\end{center}
\vspace{-10pt}
   \caption{The testing flow of our proposed framework CI-GAN. We first obtain the initial $\w'$ based on the target task, where we randomly sample a $\w'$ for the text-to-image generation task and use the inverted latent code from the original image for the text-based image manipulation task. The initial $\w'$ and the extracted text feature $\txt$ are fed into the latent space alignment model, where $\w'$ are optimized towards the direction of $\txt$. The optimized $\w'$ will be fed into the StyleGAN generator to generate image.}
\vspace{-15pt}
\label{fig:test}
\end{figure}

In the inference phase, we show the demonstration in Figure \ref{fig:test}. For the text-to-image generation task, we randomly sample a $\w'$ as the initial $\w'$, based on which the GAN model can only generate an image with random semantic attributes. Hence, we input the initial $\w'$ and the extracted text features $\txt$ from trained $E_{text}(\cdot)$ to the latent space alignment model, the $\w'$ will be optimized towards the direction of $\txt$. For the text-based image manipulation task, we use the inverted latent codes from the original real images as the initial $\w'$. When we optimize $\w'$, we apply an extra perceptual loss between the original and generated images, to preserve some text-irrelevant attributes on the generated images.

We give more details in the following sections.

\subsection{GAN Inversion Encoder}
The objective of a typical GAN model is to generate high-quality images from the random noise input, which can be denoted as $G(\cdot): \Z\rightarrow\X$. GAN inversion studies mapping the generated image back to the latent code $\z'$, such that $\z'$ can be close to the original input $\z$ and the reconstructed image from  $\z'$ can be the same as the real image. 

The recent proposed StyleGAN \cite{karras2019style,karras2020analyzing} uses Multi-Layer Perceptron (MLP) to map the initial latent space $\Z$ to another latent space $\W$, where the mapped latent codes $\w \in \W$ are adopted to synthesis images. It has been proven that the model can learn more disentangled semantics in the space $\W$ \cite{shen2020interpreting,karras2019style}, since we aim to discover the semantics of the latent codes, we choose the space $\W$ of StyleGAN2 \cite{karras2020analyzing} to conduct the experiment analysis.

For the GAN inversion task, training an encoder $E(\cdot)$ is a practical solution \cite{zhu2020indomain,xia2020tedigan}, which takes the images as the input, and outputs $\w'$ aligned with the original $\w$. To allow the reconstructed image from $\w'$ to have the same semantic information as the original image, we train the encoder $E(\cdot)$ with the pixel losses and the perceptual loss \cite{johnson2016perceptual}. Note that we use the real images $\x$ as inputs for $E(\cdot)$ training, in an attempt to make the encoder more applicable to real image manipulation. Technically, the perceptual loss regularizes the semantics of the reconstructed images by aligning extracted VGG \cite{simonyan2014very} features between the real images $\x$ and reconstructed images $\x'=G(E(\x))$. The pixel loss and perceptual loss can be formulated as
\begin{align}
  &\begin{aligned}
    \Loss_{pix} = ||\x - \x'||_2,
  \end{aligned} \\
  &\begin{aligned}
    \Loss_{vgg} = ||F(\x) - F(\x')||_2,
  \end{aligned}
\end{align}
where $||\cdot||_2$ denotes the $l_2$ distance, $F(\cdot)$ and $E(\cdot)$ denote the VGG feature extraction model and GAN inversion encoder respectively.

\subsection{Cycle-Consistent Constraint}
However, the pixel loss and perceptual loss only give regularization on the image domain, we also need to apply consistent constraints on the latent space domain, such that the inverted latent code $\w'$ can be matched with $\w$ and the latent codes can be manipulated and interpretable. Specifically, we propose to use cycle consistency training on the GAN inversion learning procedure. We adopt the adversarial loss on both the inverted code $\w'$ and the generated images from $\w'$.

The adversarial loss on the generated images can ensure the reconstructed images $\x'$ to be realistic enough, where we involve the discriminator to compete with the encoder $E(\cdot)$. The adversarial training method also helps push the inverted code $\w'$ to better fit the semantic knowledge of the generator, which can be denoted as
\begin{equation}
\begin{aligned}
\Loss_{adv}^\x = \underset{\x\sim P_{data}}\E[D(\x')],
\end{aligned} 
\end{equation}
where $D(\cdot)$ denotes the discriminator of GAN.

Moreover, we introduce to apply the adversarial loss on the the inverted codes, making the original latent code $\w$ and the inverted latent code $\w'=E(G(\w))$ to follow the same distribution. As a result, this helps us modify $\w'$ and obtain images with our desired attributes better. This can be denoted as
\begin{equation}
\begin{aligned}
\Loss_{adv}^\w = \underset{\w\sim P_{\w}}\E[D_{\w}(\w')],
\end{aligned} 
\end{equation}

We also enforce the $\w$ and inverted codes $\w'=E(G(\w))$ to be identical. The whole training objective of GAN inversion encoder can be given as
\begin{align}
  &\begin{aligned}
    \min_{\Theta_E}\Loss_E =  &\Loss_{pix} \
                              +\lambda_{vgg} \Loss_{vgg} \
                              -\lambda_{adv}^\x \Loss_{adv}^\x\\
                              &+\lambda_{\w}||\w - \w'||_2\
                              -\lambda^{\w}_{adv} \Loss_{adv}^\w, \label{eq:encoder}
  \end{aligned} \\
  &\begin{aligned}
    \min_{\Theta_D}\Loss_D = &\Loss_{adv}^\x
                              -\underset{{\x\sim P_{data}}}\E[D(\x)], \label{eq:discriminator}
  \end{aligned} \\
  &\begin{aligned}
    \min_{\Theta_{D_\w}}\Loss_{D_\w} = &\Loss_{adv}^\w
                              -\underset{\w\sim P_{\w}}\E[D_{\w}(\w)],
  \end{aligned} \label{eq:w_discriminator}
\end{align}
where $\Theta$ denotes the parameters of models and $D_w(\cdot)$ denotes the discriminator of GAN inverse codes. $\lambda_{vgg}$, $\lambda_{adv}^\x$, $\lambda_{\w}$ and $\lambda^{\w}_{adv}$ are the trade-off loss weights.

\subsection{Text Encoder by Latent Space Alignment}
With the learned GAN inversion model, we can get the initial inverted latent code $\w'$ for each real image. To control the generated images based on the text descriptions, we use the latent space alignment and discover the relationships between the explicit text semantics and the implicit space $\W$. Specifically, we adopt the paired text descriptions and $\w'$ converted from the corresponding images for similarity learning. We first use a one-layer LSTM $E_{text}(\cdot)$ to obtain the features $\txt$ of the given text descriptions, whose dimension is the same as the dimension of $\w'$. Here our goal is to learn the text feature representations $\txt$ and map them to the same feature space as space $\W$, since the StyleGAN model can learn disentangled space $\W$, aligning $\txt$ with $\w$ makes the text encoder better capture the useful semantics in the text descriptions. 

We take the InfoNCE loss \cite{oord2018representation} to learn the similarity relationships for $(\txt, \w')$. The original InfoNCE loss is used for classification task and optimizes the model through the class probabilities, it is hard for us to directly use the InfoNCE loss, as we try to uncover the pair wise relationships. Hence we integrate the $l_2$ distance into the InfoNCE loss, which is formulated as
\begin{align}
\Loss_{sim} &= - \mathop{{}\mathbb{E}}_T\left[\log \frac{exp(||\txt_{i+k} - \w'_i||_2)}{\sum_{t_j \in T} exp(||t_j - \w'_i||_2)}\right], \label{loss}
\end{align}
where $T$ represents all the text representations in the mini-batch, $i$ denotes one sample index in the training batch, and index $k$ denotes a non-zero number. $\txt_{i+k}$ represents the unpaired samples to the $\w’_i$. We aim to minimize the $l_2$ distance between paired $\txt$ and $\w'$, while maximize the distances between unpaired instances. The reason that we do not choose pair-wise loss functions, such as cosine loss, to optimize the similarity model is that pair-wise loss only considers one pair for each sample in the iterations, which fails to formulate the relationships between pairs. While in the InfoNCE loss, we can sample all the positive and negative pairs and optimize them together, which yields better matching performance. 
 
\subsection{Inference with Text-Guided Optimization}

In this module, we optimize the initial inverted latent codes $\w'$ for the text-to-image generation and text-based image manipulation tasks. To interpret the semantic attributes in the given text descriptions, we push $\w'$ towards the direction of text representations $\txt$. For the image manipulation task, we apply an extra perceptual loss \cite{johnson2016perceptual} between the original images and the generated images from the optimized $\w'$, to preserve the text-irrelevant semantic attributes of the original shapes.  

The optimization process can be given as
\begin{align}
  \begin{aligned}
    \w'_{opt} = \arg\min_{\w'}\ ||\txt -\w'||_2\ &+ ||F(\x) - F(G(\w'))||_2.
  \end{aligned} \label{eq:optimization}
\end{align}
where $\x$ denotes the original images. Note that the latter term, which is the perceptual loss, will be calculated on the image manipulation task, while for the text-to-image generation task, we only use the $l_2$ loss.

\section{Experiments}

\subsection{Datasets and Evaluation Metrics}
We evaluate our proposed framework with two challenging datasets: Recipe1M \cite{salvador2017learning} and CUB \cite{wah2011caltech} datasets. Recipe1M dataset is a public food dataset with large diversity, which has complex and fine-grained details on various classes of food images. There are $238,999$ image-recipe pairs for training, $51,119$ pairs for validation and $51,303$ pairs for testing. We use the paired ingredients only to generate the food images. CUB dataset is widely used for text-to-image generation task, which contains $200$ classes and $11,788$ images in total. There are $8,855$ and $2,933$ images for training and testing respectively. For each bird image in CUB dataset, there are $10$ English text descriptions. 

We use the quantitative evaluation metrics of Inception Score (IS) \cite{salimans2016improved} and Fr\'echet Inception Distance (FID) \cite{heusel2017gans}. To be specific, the IS computes the Kullback-Leibler (KL) divergence between conditional distribution and the marginal distribution of predicted image labels by the pretrained Inception-V3 network \cite{simonyan2014very}. Higher IS indicates the model can generate more diverse and realistic images. However, IS may fail to reflect the generated image quality in some text-to-image cases \cite{li2019object}. Therefore, we also use the FID for the evaluation, which is more robust and aligns human qualitative evaluation \cite{tao2020df}. The FID computes the Fr\'echet Inception Distance between the distributions of real and generated images in the feature space of pretrained Inception-V3 network \cite{simonyan2014very}. 

\subsection{Implementation Details}
Our proposed unified framework is implemented based on StyleGAN2-Ada \cite{Karras2020ada}. As is shown in Table \ref{table:ablation}, we find our proposed CI-GAN can get the best IS and FID compared with the conventional text feature-conditioned StyleGAN2 model. We also empirically observe the label-conditioned StyleGAN2 can obtain better performance than unconditional StyleGAN2. We optimize the model with $\lambda_{vgg}=0.00005$, $\lambda_{adv}^\x=0.08$, $\lambda_{\w}=0.01$ and $\lambda_{adv}^\w=0.005$, where we follow the settings in \cite{zhu2020indomain} to set part of the loss weights. We follow \cite{zhu2020indomain} to build the GAN inversion encoder, consisting of $8$ ResNet blocks. In Recipe1M dataset, we only take ingredient as input to generate food images, as intuitively ingredients have provided rich semantic attributes. In CUB dataset, we randomly sample one descriptive sentence for each image when training. Note that when we do text-based image manipulation task, we first convert the original image to the latent code $\w'$ and optimize $\w'$ with Equation (\ref{eq:optimization}). While when we do the text-to-image generation task, we first randomly sample a latent code $\w'$ and optimize the $\w'$ with Equation (\ref{eq:optimization}) without using the perceptual loss. 

\begin{table}
\centering
  \caption{We show result comparison on CUB dataset between the conventional text feature-conditioned StyleGAN2 model and our proposed framework.}
\vspace{-10pt}
\label{table:ablation}
\begin{tabular}{l|cc}
\toprule
GAN Structure      & Inception Score↑ & FID↓ \\
\midrule
Text-Conditioned StyleGAN2 & & \\ 
\quad\quad - Plain LSTM         & 4.37 ± 0.05    & 14.07  \\
\quad\quad - Pretrained BERT \cite{devlin2018bert} Model & 5.43 ± 0.04  & 10.67 \\
\hline
CI-GAN (Ours)   &  \textbf{5.72 ± 0.12}     & \textbf{9.78} \\
\bottomrule 
\end{tabular}
\vspace{-10pt}
\end{table}

\begin{table}
\centering
  \caption{The inception score and FID of our proposed method against various models on Recipe1M dataset.}
\vspace{-10pt}
\label{table:food}
\begin{tabular}{l|cc}
\toprule
Method     & Inception Score↑ & FID↓ \\
\midrule
$\mathbf{R^2 GAN}$ \cite{zhu2019r2gan}      & 4.54 ± 0.07 & -      \\
StackGAN++ \cite{zhang2018stackgan++} & 5.03 ± 0.09   & -    \\
CookGAN  \cite{zhu2020cookgan}   & 5.41 ± 0.11    &  -   \\
CI-GAN (Ours)        & \textbf{5.97 ± 0.11} & \textbf{9.12} \\
\bottomrule          
\end{tabular}
\vspace{-10pt}
\end{table}

\begin{table}
\centering
  \caption{The inception score and FID of our proposed method compared with various models on CUB dataset.}
\vspace{-10pt}
\label{table:cub}
\begin{tabular}{l|cc}
\toprule
Methods    & Inception Score↑ & FID↓  \\
\midrule
StackGAN++ \cite{zhang2018stackgan++} & 4.04 ± 0.06     & -     \\
AttnGAN \cite{xu2018attngan} & 4.36 ± 0.03     & 23.98 \\
MirrorGAN \cite{qiao2019mirrorgan}  & 4.56 ± 0.05     & -     \\
ControlGAN \cite{li2019controllable} & 4.58 ± 0.09 & - \\
SD-GAN  \cite{yin2019semantics}  & 4.67 ± 0.09      & -     \\
DM-GAN \cite{zhu2019dm}   & 4.75 ± 0.07     & 16.09 \\
DF-GAN \cite{tao2020df}    & 4.86 ± 0.04     & 19.24 \\
RiFeGAN \cite{cheng2020rifegan} & 5.23 ± 0.09 & - \\
DALL-E (zero-shot) \cite{ramesh2021zero} & 2.85 ± 0.04  & 51.00 \\
CI-GAN (Ours)       & \textbf{5.72 ± 0.12}     & \textbf{9.78} \\
\bottomrule 
\end{tabular}
\vspace{-10pt}
\end{table}

\begin{figure*}
\begin{center}
\includegraphics[width=0.8\textwidth]{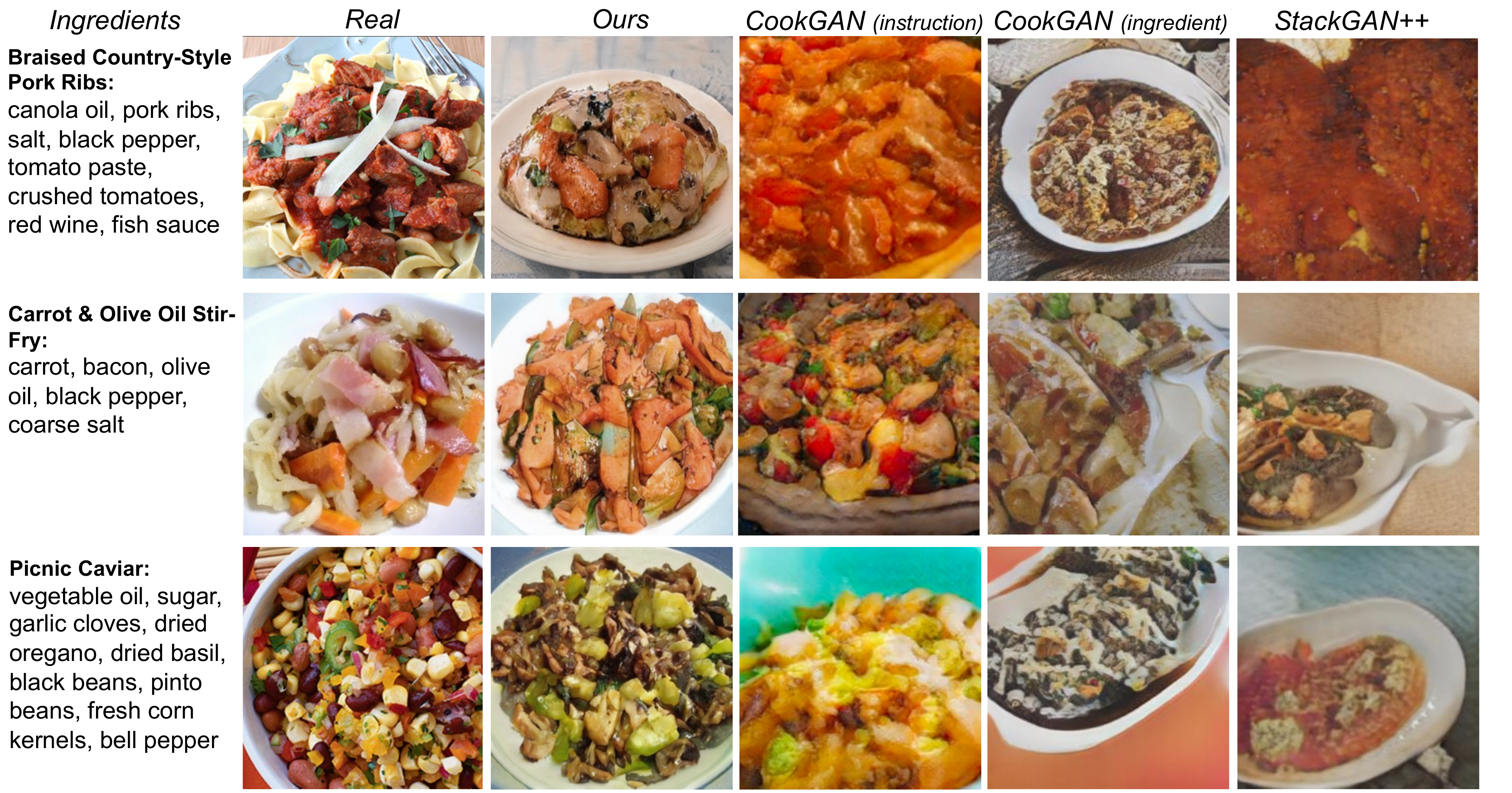}
\end{center}
\vspace{-10pt}
   \caption{The generated food images based on the ingredients except \emph{CookGAN (instruction)}. We show the visualizations of real images, our proposed framework, CookGAN using cooking instructions or ingredients, and StackGAN++.}
\vspace{-10pt}
\label{fig:food_comp}
\end{figure*}

\begin{figure*}
\begin{center}
\includegraphics[width=0.8\textwidth]{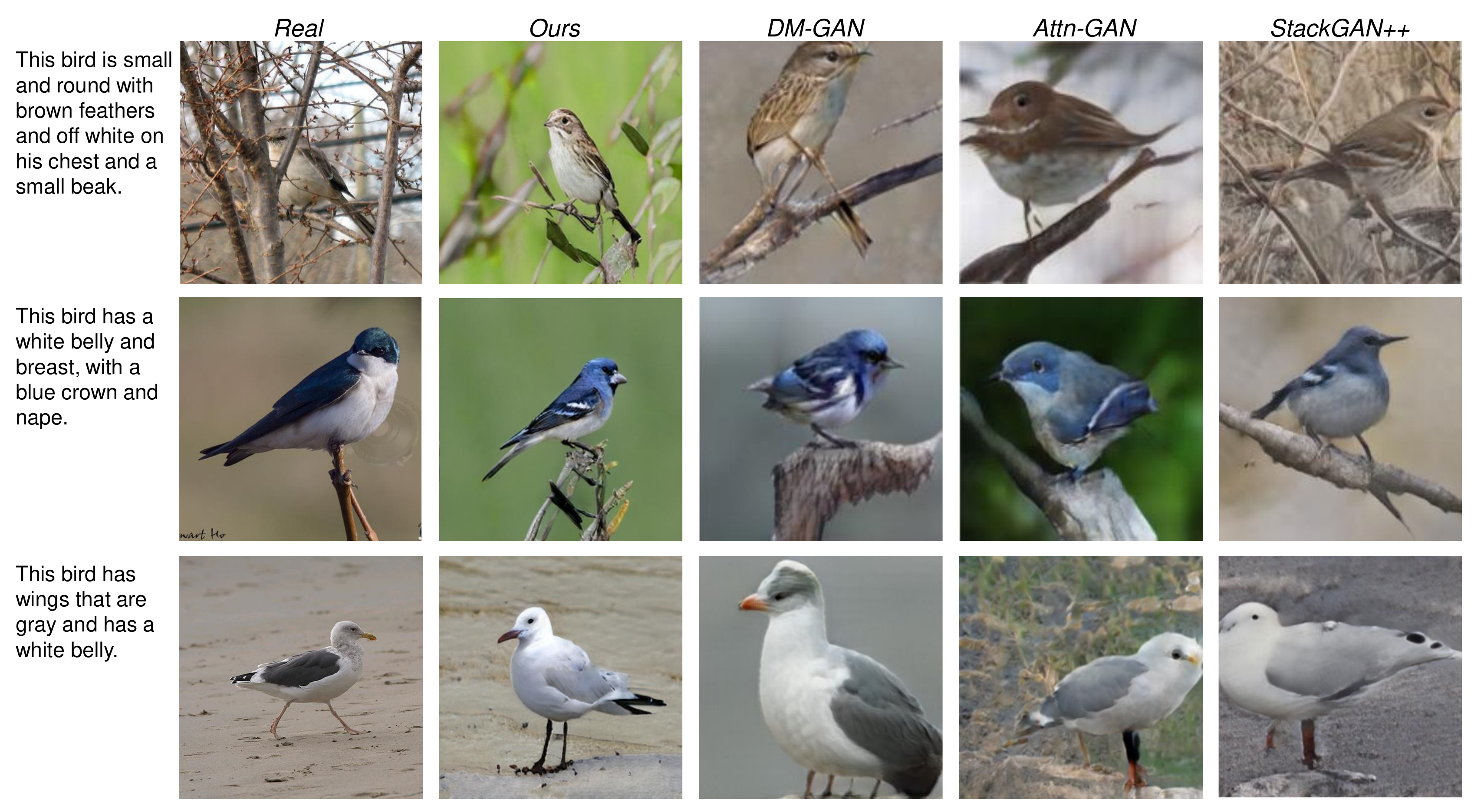}
\end{center}
\vspace{-10pt}
   \caption{The text-to-image generation based on the text descriptions at the left side. We show the visualizations of real images, our proposed framework, DM-GAN, Attn-GAN and StackGAN++.}
\vspace{-10pt}
\label{fig:bird_comp}
\end{figure*}

\subsection{Quantitative Results}
We compare our proposed framework with various state-of-the-art text-to-image synthesis frameworks. 

We show the quantitative evaluation results on Recipe1M dataset in Table \ref{table:food}. $\mathbf{R^2 GAN}$ \cite{zhu2019r2gan} is a naive baseline to generate low-resolution images of $64 \times 64$, which are further resized to $256 \times 256$ by linear interpolation. StackGAN++ \cite{zhang2018stackgan++} introduces multiple generators and discriminators for high-resolution image generation. CookGAN \cite{zhu2020cookgan} is the state-of-the-art text-to-image generation framework in Recipe1M dataset, which is an extension based on the StackGAN++ architecture. The difference between CookGAN and StackGAN++ mainly lies in that CookGAN adopts the attention mechanism to improve the conditional recipe features. Our method outperforms all previous methods by a large margin, specifically we obtain over $10 \%$ and $30\%$ improvement over CookGAN in the IS and FID respectively. This illustrates the proposed CI-GAN framework can synthesize more realistic food images with better text-image semantic consistency.

When we shift to the CUB dataset, we show the results in Table \ref{table:cub}. The CUB dataset has been widely used for the general text-to-image generation task. It can also be observed that our proposed CI-GAN can achieve the best results across the IS and FID, which are $5.68$ and $5.41$ respectively. Most of the previous works, such as AttnGAN \cite{xu2018attngan}, MirrorGAN \cite{qiao2019mirrorgan} and RiFeGAN \cite{cheng2020rifegan}, use the conditional GAN structure equipped with the attention mechanism to improve the GAN representation and further boost the generation performance. DALL-E \cite{ramesh2021zero} is the newly proposed transformer-based text-to-image generator, they use external large-scale datasets to train the generation model. Since they use the zero-shot learning, where they did not train their model on the CUB dataset, hence their results are relatively poor. While our method can outperform existing results using a novel architecture with the idea of inverse GAN, which is effective for text-to-image generation task. Specifically, we train our model without using external large-scale datasets, we learn the similarity between the latent codes and the text representations in the latent space alignment model with a plain LSTM only. It suggests the efficacy of our adapted InfoNCE loss and the disentanglement of our learned text representations. 

\begin{figure}
\begin{center}
\includegraphics[width=0.48\textwidth]{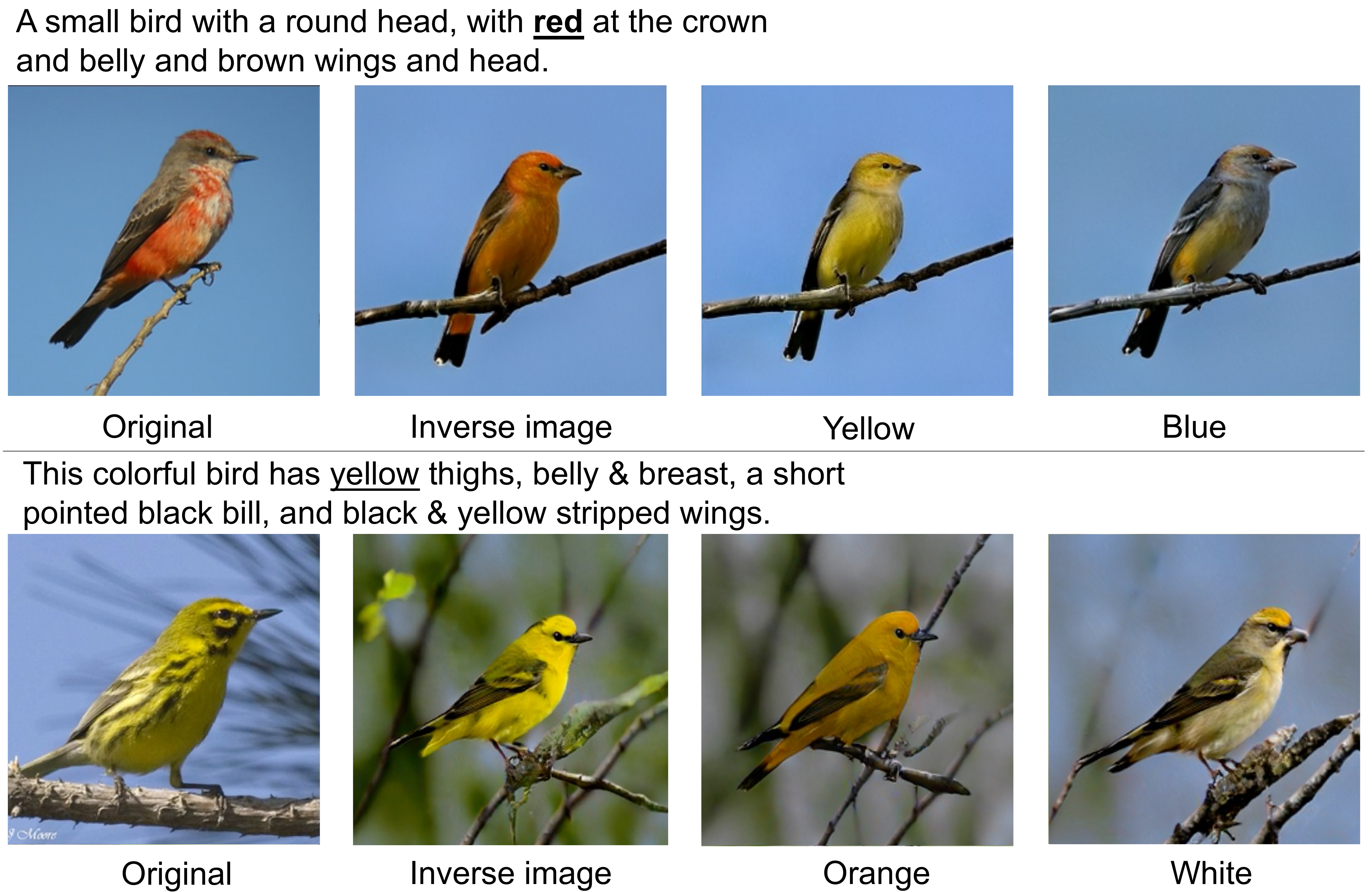}
\end{center}
\vspace{-10pt}
   \caption{The manipulation of given bird images. Words with underline indicate the manipulated words. We replace these words with different colour attributes and show the corresponding generated results.}
\vspace{-15pt}
\label{fig:bird_mani}
\end{figure}

\subsection{Qualitative Results}
\textbf{Text-to-image Generation.} The visualizations of generated images on Recipe1M and CUB datasets are shown in Figure \ref{fig:food_comp} and \ref{fig:bird_comp} respectively, where part of the generated images are taken from \cite{tao2020df} and \cite{zhu2020cookgan}. Food image generation requires detailed interpretation of complex ingredients. In Figure \ref{fig:food_comp}, we show generation results for three kinds of food. We show the CookGAN results using cooking instructions and ingredients as conditional input respectively. Our method only uses the ingredient information as the input textual descriptions, while the images generated by our proposed framework can produce most text-relevant visual contents and have high quality. For example in the second row, where the input ingredients contain \emph{carrot}, our generated images have plausible items that have similar shapes and colours to \emph{carrot}. However, it is hard to generate all the ingredient pellets clearly, which can be observed in the last row, our model can hardly generate various ingredients except the \emph{bell pepper} given the text input. Generally, our proposed framework can generate food images with more matched semantic attributes with the ingredients than other methods. In Figure \ref{fig:bird_comp}, we show results in CUB dataset. We can see that our proposed method not only can generate images corresponded with the given text input, but also can synthesis realistic images with high diversity. In contrast, many existing works are limited by the generation quality and incorrect semantic interpretation. 

\noindent \textbf{Image Manipulation.} In Figure \ref{fig:bird_mani}, we show some manipulated results in CUB dataset. Specifically, we mainly replace some colour attributes, which can be easily observed in our manipulated images. For example in the top row, we change the colour of the \emph{crown} and \emph{belly}, while maintain the \emph{brown wings}. We can see that only the colour at the \emph{belly} has been changed in the manipulated results, while the colour of \emph{wings} remains as brown. However, it is hard to get perfect semantic attribute preservation, such as shape and pose, after manipulation, as in the bottom row. There are two main reasons. One reason is that we cannot get the inversion results that are the same as the original images, then in the $\w'$ optimization process the model loses part of the original semantic information. Another reason is that the latent space $\W$ of StyleGAN is still entangled with some semantic attributes \cite{xia2020tedigan}, which means that when we change part of the attributes, the rest attributes will also be affected. We leave improving the space $\W$ with better disentanglement for the future work.

\begin{figure}
\begin{center}
\includegraphics[width=0.48\textwidth]{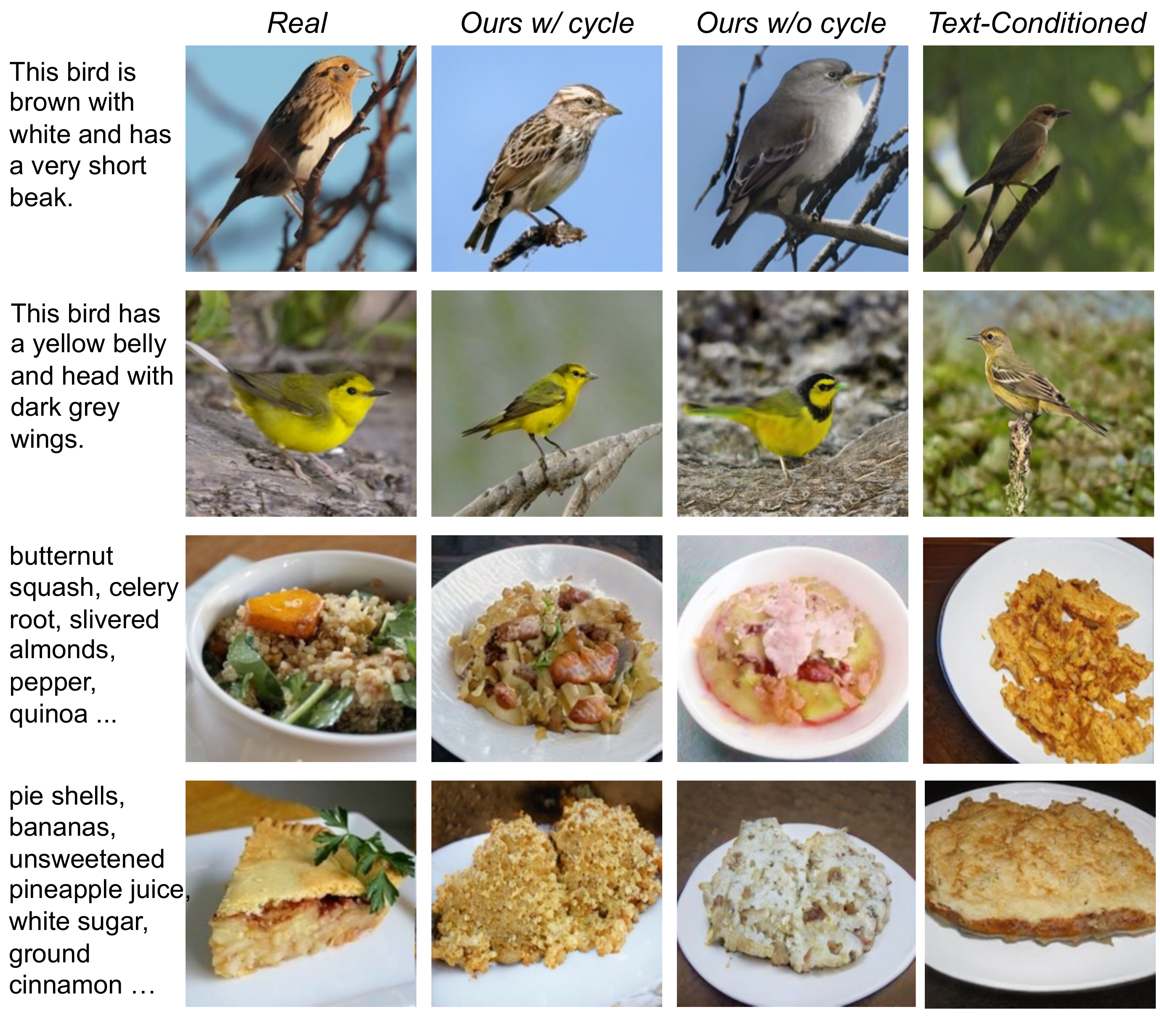}
\end{center}
\vspace{-10pt}
  \caption{The ablation studies. We show images generated by our proposed framework with and without cycle consistent training, and pretrained text feature-conditioned model. The model trained with cycle consistency can generate the most text-relevant images.}
\vspace{-10pt}
\label{fig:ablation}
\end{figure}

\begin{figure}
\begin{center}
\includegraphics[width=0.48\textwidth]{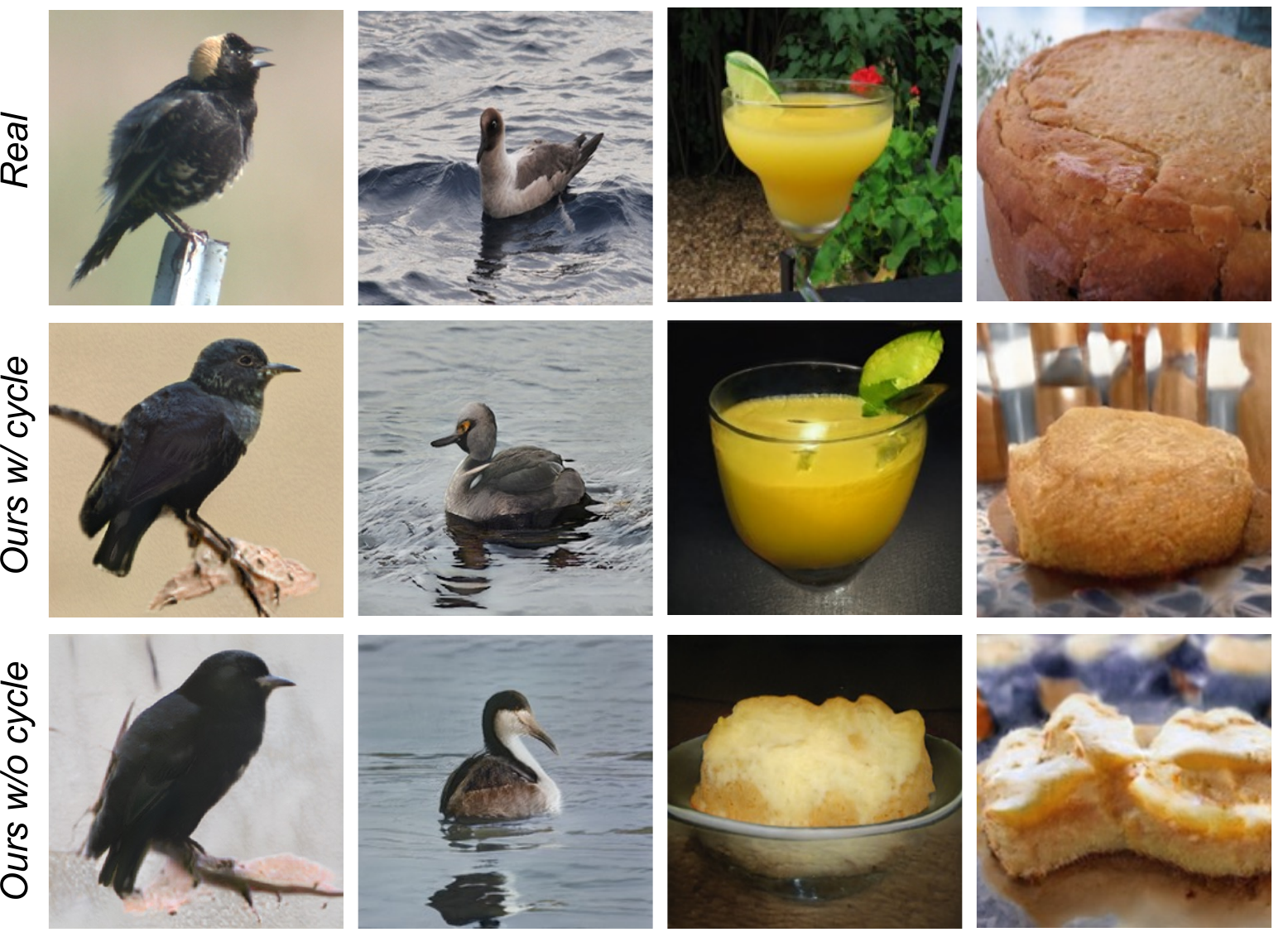}
\end{center}
\vspace{-10pt}
  \caption{The GAN inversion results. For different real images, we show the inversion results from the GAN inversion models trained with and without cycle-consistency loss. GAN inversion encoder trained with cycle consistency can preserve more original image properties than that trained without cycle consistency.}
\vspace{-12pt}
\label{fig:inverse_demo}
\end{figure}

\subsection{Ablation Studies}
\textbf{The comparison between text-conditioned GAN and CI-GAN.} In Table \ref{table:ablation}, we compare the text-to-image generation performance between conventional text-conditioned GAN and our proposed CI-GAN on CUB dataset. Specifically, we experiment with different GAN conditional inputs of LSTM text features and pretrained BERT \cite{devlin2018bert} features. We can find that the conditional GAN with direct text features as input does not have the best results on the IS and FID, while the proposed CI-GAN can get better results. It indicates that using text features as input will affect the GAN generation performance, since the limited paired text-image combinations will limit the diversity of generated images. We also show the qualitative comparison in Figure \ref{fig:ablation}. The generated images by our method are better aligned with the textual descriptions, while the generation by the text feature-conditioned GAN model may not have enough diversity. Moreover, some generated images from the text-conditioned GAN, e.g., images in the third row in Figure \ref{fig:ablation}, fail to match with the text inputs correctly.

\begin{figure}
\begin{center}
\includegraphics[width=0.5\textwidth]{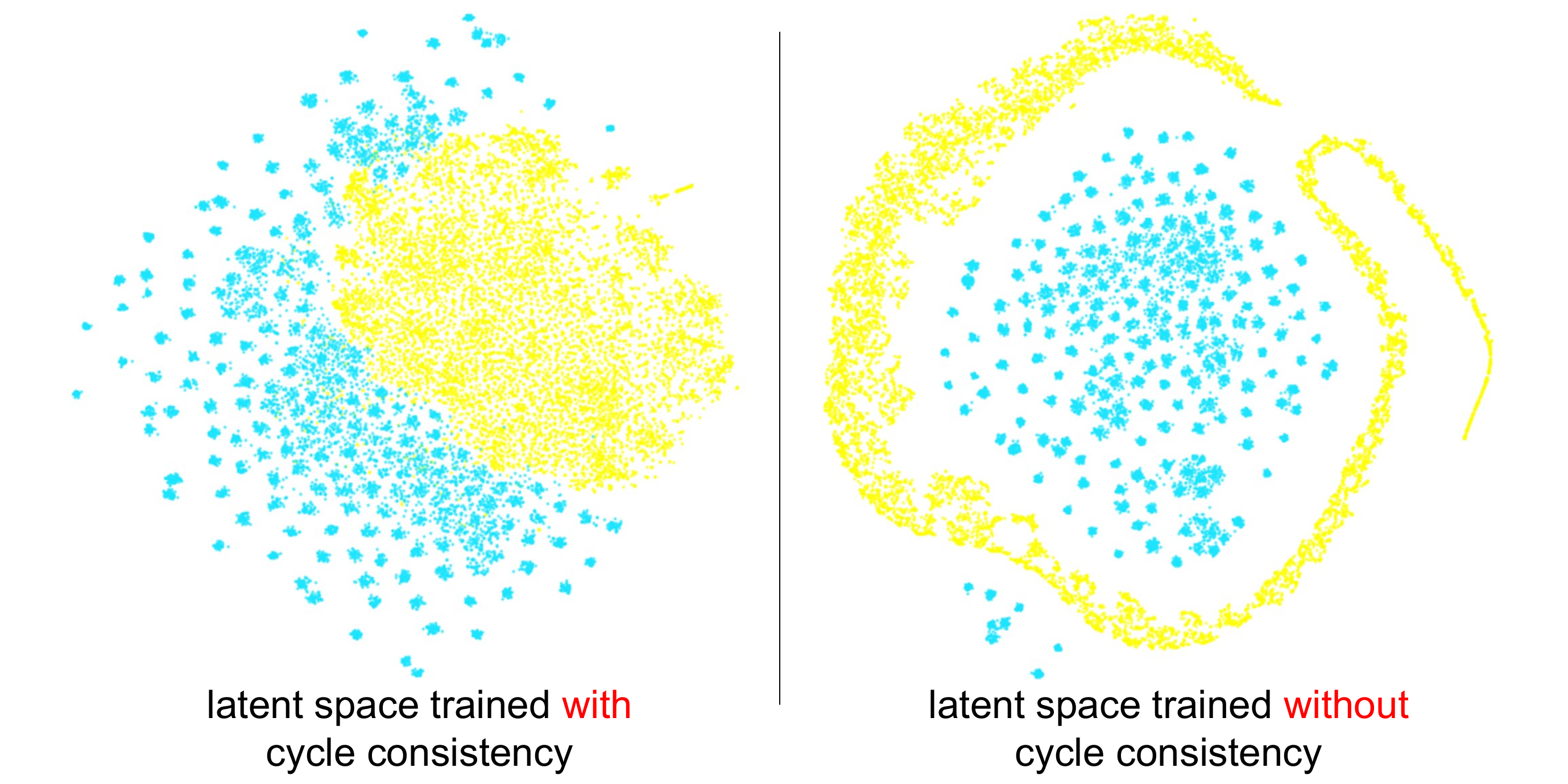}
\end{center}
\vspace{-10pt}
   \caption{The latent space visualization of $\w$ trained with and without cycle consistency in the CUB dataset. The blue and yellow dots denote the original latent codes and learned inverted codes respectively.}
\vspace{-15pt}
\label{fig:cyc_enc}
\end{figure}

\noindent \textbf{GAN inversion models with and without cycle-consistency training.} To compare the difference between our proposed frameworks with and without the cycle-consistency training, we first show the GAN inversion results in Figure \ref{fig:inverse_demo}. We find the GAN inversion model trained without the cycle-consistency loss may fail to preserve the original semantic attributes (e.g. the third column) and the object pose information (e.g. the second column). We further analyse the distribution of the latent codes by t-SNE \cite{van2008visualizing}, which is shown in Figure \ref{fig:cyc_enc}. 
The inverted latent codes with cycle-consistency training have overlaps and similar distributions with the original latent codes. While the distribution of the inverted codes without cycle-consistency training fail to match that of the original latent codes correctly. Such misalignment further affects the performance of text-to-image generation, since the latent space alignment model is trained based on the learned inverted latent codes. For instance, in the first row of Figure \ref{fig:ablation}, the generated bird image from model without cycle-consistency training fails to interpret the \emph{brown} colour attributes of the given textual descriptions.

\section{Discussion}
The text-conditioned GAN architecture has been widely used in the text-to-image generation task. We have researched on the impacts of different GAN structures in Table \ref{table:ablation}, and find that using paired text-image data for GAN training reduces the generation quality and diversity. While the proposed CI-GAN without paired text feature inputs can yield better results. With the well-trained StyleGAN, we can update inverted latent codes $\w'$ in the latent space $\W$ to get desired outputs. However, the latent space of the adopted StyleGAN may not be disentangled enough with various semantic attributes, such that the model can hardly only change one attribute of the generated image while keep other text-irrelevant attributes. This leaves some room for the future improvement. 

Another benefit of using StyleGAN as the backbone network for text-to-image generation is that, StyleGAN has good capability to generate high-resolution images. Therefore, in the future we can easily extend our proposed framework to generate higher resolution images, such as $1024\times1024$. Here we only generate images of size $256\times256$ to compare with the existing works.

\section{Conclusion}
In this paper, we have proposed a novel unified framework of Cycle-consistent Inverse GAN (CI-GAN) for the text-to-image generation and text-guided image manipulation tasks, where we are the first to incorporate the idea of GAN inversion into the image generation task on two challenging datasets in the wild. To be specific, we first train a StyleGAN model without conditional text input, such that the model can produce images with high diversity and good quality. Then we learn a GAN inversion model with our proposed cycle consistency training, to convert the images back to the GAN latent space and obtain the inverted latent codes. To generate images with our desired semantic attributes, we further discover the semantics of the GAN latent space. We learn a similarity model between text representations and the latent codes. The learned similarity model is adopted to optimize the latent codes for the text-to-image generation and image manipulation tasks. We have conducted extensive experiments and analysis on the Recipe1M and CUB datasets, and illustrate the superior performance of our proposed framework.

\section*{Acknowledgement}
This research is supported, in part, by the National Research Foundation (NRF), Singapore under its AI Singapore Programme (AISG Award No: AISG-GC-2019-003) and under its NRF Investigatorship Programme (NRFI Award No. NRF-NRFI05-2019-0002). Any opinions, findings and conclusions or recommendations expressed in this material are those of the authors and do not reflect the views of National Research Foundation, Singapore. This research is also supported, in part, by the Singapore Ministry of Health under its National Innovation Challenge on Active and Confident Ageing (NIC Project No. MOH/NIC/COG04/2017 and MOH/NIC/HAIG03/2017), and the MOE Tier-1 research grants: RG28/18 (S) and RG22/19 (S).

\bibliographystyle{ACM-Reference-Format}
\bibliography{sample-base}

\end{document}